# TOWARDS RELIABLE SUBSEA OBJECT RECOVERY: A SIMULATION STUDY OF AN AUV WITH A SUCTION-ACTUATED END EFFECTOR

Michele Grimaldi[1,2,*], Yosaku Maeda[2], Hitoshi Kakami[2], Ignacio Carlucho[1], Yvan Petillot[1], Tomoya Inoue[2]

[1]Heriot-Watt University, Edinburgh, UK
[2]Japan Agency for Marine-Earth Science and Technology (JAMSTEC), Yokosuka, Japan

**ABSTRACT**

*Autonomous object recovery in the hadal zone is challenging due to extreme hydrostatic pressure, limited visibility and currents, and the need for precise manipulation at full ocean depth. Field experimentation in such environments is costly, high-risk, and constrained by limited vehicle availability, making early validation of autonomous behaviors difficult. This paper presents a simulation-based study of a complete autonomous subsea object recovery mission using a Hadal Small Vehicle (HSV) equipped with a three-degree-of-freedom robotic arm and a suction-actuated end effector. The Stonefish simulator is used to model realistic vehicle dynamics, hydrodynamic disturbances, sensing, and interaction with a target object under hadal-like conditions. The control framework combines a world-frame PID controller for vehicle navigation and stabilization with an inverse-kinematics-based manipulator controller augmented by acceleration feed-forward, enabling coordinated vehicle–manipulator operation. In simulation, the HSV autonomously descends from the sea surface to 6,000 m, performs structured seafloor coverage, detects a target object, and executes a suction-based recovery. The results demonstrate that high-fidelity simulation provides an effective and low-risk means of evaluating autonomous deep-sea intervention behaviors prior to field deployment.*

## 1. INTRODUCTION

The exploration and scientific study of the hadal zone, defined as ocean depths exceeding 6,000 m, presents extreme technical and operational challenges. Hydrostatic pressures surpassing 60 MPa, near-total darkness, low temperatures, and complex seafloor terrain significantly complicate both vehicle operation and manipulation tasks. At the same time, scientific interest in hadal environments has grown rapidly in recent years, driven by advances in deep-sea technology and an increasing recognition of the ecological and geophysical importance of the deepest ocean trenches [1]. This growth has intensified the demand for reliable robotic systems capable of autonomous operation and physical interaction at full ocean depth. While deep-sea robotics has enabled unprecedented access to these environments, autonomous object recovery remains particularly difficult due to the tight coupling between vehicle stabilization, perception, and dexterous manipulation. Field experimentation at hadal depths is inherently expensive and high-risk. The deployment of fully ocean-depth vehicles requires specialized vessels, limited operational windows, and extensive safety margins, often restricting testing to a small number of missions. As a consequence, validating control strategies, autonomy pipelines, and vehicle–manipulator coordination directly in the field is impractical during the early development stages. Failures at depth may result in irreversible system loss, further emphasizing the need for robust pre-deployment validation. High-fidelity simulation offers a practical alternative for developing and evaluating autonomous deep-sea robotic systems. To the best of our knowledge, no prior work has simulated autonomous missions at hadal depths, with most studies stopping at shallow- or mid-water environments. In this work, we leverage the Stonefish marine robotics simulator, which has been enhanced to model depth-dependent hydrodynamics, mass distribution shifts, and volume compression due to extreme pressure. These effects also alter onboard sensors, such as IMUs, affecting state estimation and vehicle control. By capturing these coupled interactions between vehicle physics, sensors, and manipulator dynamics, Stonefish enables realistic, system-level testing that was previously infeasible. Using this improved simulator, we model a complete autonomous object recovery mission for the Hadal Small Vehicle (HSV) [2] equipped with a three-degree-of-freedom robotic arm and a suction-based end effector [3]. The vehicle is controlled using a world-frame PID controller informed by simulated inertial navigation data, while the manipulator employs inverse kinematics with acceleration feed-forward to achieve precise and stable motion during interaction tasks. The simulated mission includes autonomous descent from the sea surface to 6,000 m, structured seafloor coverage, visual detection of a target object, and coordinated vehicle–manipulator control to execute a suction-based pickup. Rather than focusing on perception algorithm performance, this study emphasizes system-level behavior and control integration, demonstrating how complex autonomous intervention behaviors can be validated entirely in simulation. This approach reduces development risk, lowers operational costs, and provides a foundation for future digital twin integration and full-ocean-depth autonomous manipulation.

---

*Corresponding author
Documentation for asmeconf.cls: Version 1.45, January 3, 2026.

## 2. RELATED WORK

Research in underwater robotics encompasses a diverse array of vehicles, manipulators, and simulation tools designed for operation in extreme marine environments. Conventional ROVs and AUVs have been extensively deployed for shallow and mid-depth exploration; however, they are generally limited in terms of operational depth, endurance, and dexterous manipulation capabilities. To access the hadal zone at depths exceeding 10,000 m, hybrid vehicles such as *Nereus* [4] were developed to combine autonomous navigation with remotely operated control, enabling scientific sampling in the deepest ocean trenches. More recently, China's *Haidou* autonomous and remotely operated vehicle has demonstrated repeated full-ocean-depth deployments during scientific expeditions in the Mariana Trench, validating the feasibility of long-duration hadal operations with hybrid control architectures [5]. Similar design philosophies are embodied in vehicles such as the Hadal ARV[1], which support both autonomous and teleoperated missions under extreme deep-sea conditions. In parallel, next-generation fully autonomous platforms such as *Orpheus* have been proposed to enable persistent hadal exploration without tethered operation, emphasizing compact design, pressure-tolerant electronics, and advanced autonomy[2]. Manipulation in subsea environments introduces significant challenges due to fluid dynamics, limited visibility, and interaction with fragile specimens. Classic work on motion and force control of underwater vehicle–manipulator systems explores dynamic coupling and control strategies for such systems [6, 7]. More recent studies focus on designing manipulators with compliant structures and innovative actuation to improve robustness and adaptability in unstructured underwater environments [8, 9]. Taxonomies of underwater manipulative actions highlight the wide range of tasks required for marine sampling, including scraping, suction, and precise grasping of biological specimens, emphasizing the need for versatile end-effector designs [10]. In addition, localization and positioning alone provide limited benefit in these environments, as visual [11] and acoustic sensing are severely degraded by low visibility, strong currents, and complex seafloor topology, making autonomous manipulation particularly challenging. Suction-based mechanisms have emerged as key tools for underwater manipulation. Recent work has demonstrated mobile manipulators equipped with suction mechanisms that maintain stable contact under hydrodynamic disturbances, providing insight into suction-assisted manipulation performance in aquatic environments [12]. Broader reviews of marine robotics for biological sampling describe a variety of ROV end-effectors, including suction samplers, and outline design requirements for effective interaction with both hard and soft substrates at depth [10]. Simulation plays a critical role in developing and validating underwater robotic systems, particularly for mission planning, perception, and control prior to field deployment [13, 14]. A range of open-source and research simulators has been developed to support autonomous underwater robotics. Frameworks such as DAVE provide generalized aquatic environments based on ROS and Gazebo, modeling vehicles, manipulators, and environmental interactions [15]. High-fidelity simulators built on game engines, such as HoloOcean, offer realistic visual rendering, multi-agent support, and advanced sensor modeling including imaging sonar and optical systems [16, 17]. More recent GPU-accelerated frameworks like OceanSim focus on accurate underwater perception and physics-based sensor rendering, but typically provide limited modeling of vehicle dynamics and fluid interactions [18]. Comparative analyses of underwater robotics simulators highlight trade-offs between physics fidelity, extensibility, and integration complexity [19]. Stonefish distinguishes itself by combining detailed hydrodynamic modeling based on rigid-body geometry with tight ROS integration and flexible scenario definition, enabling coupled vehicle–manipulator dynamics alongside realistic sensor and environmental effects. In addition to its physics capabilities, Stonefish has also been shown to support visually realistic underwater scenes suitable for validating perception pipelines in simulation [20]. This combination makes Stonefish particularly well suited for simulating complex deep-sea interaction scenarios such as autonomous object recovery in the hadal zone, where both accurate dynamics and realistic environmental interactions are essential. Finally, digital twin concepts and simulation-based frameworks further support subsea robotic research by enabling virtual testing and predictive assessment of system behavior under varying operational conditions [21, 22]. The integration of advanced simulation, robust vehicle–manipulator control, and adaptable end-effector mechanisms provides a strong foundation for enabling autonomous object recovery in extreme environments such as the hadal zone.

## 3. VEHICLE AND ARM CONTROLLERS

The Hadal Small Vehicle (HSV), showed in figure 1, is controlled using a world-frame PID controller that regulates its position along the $x$, $y$, and $z$ axes as well as its yaw orientation. Let $\mathbf{p} = [x, y, z, \psi]^\top$ denote the vehicle pose in the world frame, and $\mathbf{p}_{wp}$ the current waypoint. The position and yaw errors are defined as:

$$\mathbf{e} = \mathbf{p}_{wp} - \mathbf{p} = \begin{bmatrix} e_x \\ e_y \\ e_z \\ e_\psi \end{bmatrix}.$$

For each axis $i \in \{x, y, z, \psi\}$, the PID control law is given by:

---
[1] https://en.wikipedia.org/wiki/Hadal_ARV
[2] Woods Hole Oceanographic Institution. *Orpheus: Autonomous Underwater Vehicle*. Available online: https://www.whoi.edu/what-we-do/explore/underwater-vehicles/auvs/orpheus/

$$u_i(t) = K_{p,i} e_i(t) + K_{i,i} \int_0^t e_i(\tau)d\tau + K_{d,i} \frac{de_i}{dt},$$

where $K_{p,i}$, $K_{i,i}$, and $K_{d,i}$ are the proportional, integral, and derivative gains, respectively. Anti-windup and deadband mechanisms are incorporated to ensure smooth control near the setpoints. The resulting world-frame forces and moment, $\mathbf{F}_w = [F_x, F_y, F_z, \tau_\psi]^\top$, are transformed into the body frame using the current yaw angle $\psi$:

$$\mathbf{F}_b = \begin{bmatrix} F_x^b \\ F_y^b \\ F_z^b \\ \tau_\psi^b \end{bmatrix} = \begin{bmatrix} \cos\psi & \sin\psi & 0 & 0 \\ -\sin\psi & \cos\psi & 0 & 0 \\ 0 & 0 & 1 & 0 \\ 0 & 0 & 0 & 1 \end{bmatrix} \mathbf{F}_w.$$

These body-frame commands are then mapped to the eight thrusters via a thruster allocation matrix, producing the final thruster setpoints.

### 3.1. Manipulator Control Using Inverse Kinematics and Acceleration Feed-Forward

The HSV is equipped with a three-degree-of-freedom (3-DOF) electric manipulator terminating in a suction-based end effector. Manipulation is performed using an acceleration-level control strategy, which is well suited for deep-sea robotic arms operating under high loads and slow, precise motion requirements, as demonstrated in recent hadal-zone manipulator developments [3]. Given a desired Cartesian target position $\mathbf{x}_{target} = [x, y, z]^\top$ expressed in the vehicle frame, the corresponding joint configuration $\mathbf{q} = [\theta_1, \theta_2, \theta_3]^\top$ is computed using geometric inverse kinematics. The joint angles are obtained as

$$\theta_1 = \arctan 2(y, x), \tag{1}$$

$$\theta_3 = -\arccos\left(\frac{x^2 + y^2 + z^2 - l_1^2 - l_2^2}{2l_1 l_2}\right), \tag{2}$$

$$\theta_2 = \arctan 2\left(z, \sqrt{x^2 + y^2}\right) - \arctan 2(l_2 \sin\theta_3, l_1 + l_2 \cos\theta_3), \tag{3}$$

where $l_1$ and $l_2$ denote the lengths of the first and second arm links, respectively. To achieve smooth and accurate trajectory tracking, joint motion is regulated at the acceleration level using a feed-forward formulation. The commanded joint accelerations are computed as

$$\ddot{\mathbf{q}}_{cmd} = \ddot{\mathbf{q}}_{target} + K_v\left(\dot{\mathbf{q}}_{target} - \dot{\mathbf{q}}\right) + K_p\left(\mathbf{q}_{target} - \mathbf{q}\right), \tag{4}$$

where $\mathbf{q}_{target}$, $\dot{\mathbf{q}}_{target}$, and $\ddot{\mathbf{q}}_{target}$ denote the desired joint positions, velocities, and accelerations, respectively, and $K_p$ and $K_v$ are diagonal proportional and velocity gain matrices. The resulting acceleration commands are numerically integrated to obtain desired joint velocities and positions, which are then transmitted to the simulated joint controllers. This acceleration-based formulation provides improved smoothness and stability compared to purely position-level control, particularly during contact-sensitive operations such as suction-based grasping. By combining vehicle stabilization with acceleration-controlled manipulation, the HSV is able to precisely position the suction end effector relative to the target object, enabling reliable autonomous object recovery under hadal-like conditions.

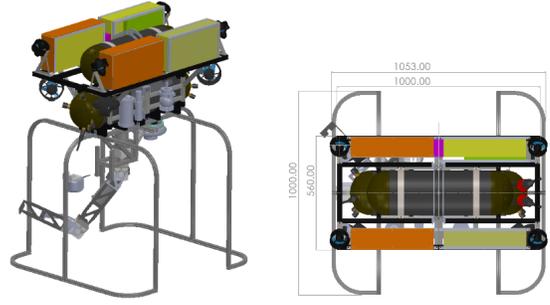

**FIGURE 1: HSV 3D model**

This scenario demonstrates the integration of vehicle stabilization, manipulator kinematics, and end-effector actuation in a fully autonomous workflow. Visual feedback from the down-looking camera, illustrated in Fig. 2, is used to detect and localize the target object during the recovery task.

## 4. EXPERIMENTS
### 4.1. Scenario

The goal of the experiment is to evaluate the autonomous object recovery workflow of the HSV in a simulated hadal environment. The scenario is structured as follows:

1. The HSV navigates autonomously toward a target starfish using the world-frame PID vehicle controller.

2. Once the vehicle approaches the starfish, the 3-DOF manipulator positions the suction cup above the target using the inverse kinematics and acceleration feed-forward controller.

3. The suction-actuated end effector performs a grasp, lifting the starfish from its resting location on the seafloor.

4. The HSV and manipulator coordinate motion to safely transport the starfish to the desired location.

This scenario demonstrates the integration of vehicle stabilization, manipulator kinematics, and end-effector actuation in a fully autonomous workflow.

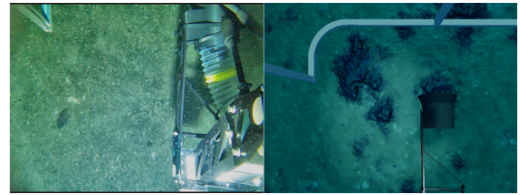

**FIGURE 2: HSV camera output comparison: (left) real-world image and (right) simulated scene camera output.**

**TABLE 1**: Simulated onboard sensors and update rates used in the experiments.

| Sensor | Rate | Usage |
|---|---|---|
| IMU | 10 Hz | Attitude and angular rate estimation |
| DVL | 10 Hz | Velocity estimation relative to seafloor |
| Pressure | 10 Hz | Depth estimation |
| INS | 10 Hz | Fused state estimate for PID control |
| Down-looking camera | 10 Hz | Starfish detection |

### 4.2. Simulation Setup

The experiment is conducted entirely in a physical high-fidelity simulation environment. The Stonefish simulator models the HSV, manipulator, and environmental interactions, including currents, buoyancy, and limited visibility typical of deep-sea conditions. A virtual starfish is placed on the seafloor as the target object. The HSV is equipped with a down-looking camera and a light, which are used for detecting the starfish, as well as a forward-looking stereo camera pair that is not utilized in the current experiments. The vehicle relies on simulated inertial navigation system (INS) feedback for closed-loop control. In the simulation, the IMU, DVL, pressure sensor, and INS provide data at 10Hz, matching the sampling rates of the real vehicle, to ensure realistic state estimation for the PID controller. The HSV begins the mission at the sea surface, dives to a depth of 6,000 meters, and navigates along a predefined path using the PID controller. When the starfish is visually detected by the down-looking camera, the vehicle stabilizes, and the manipulator executes the pick-up procedure as described above. The specific visual detection algorithm is treated as an external perception module and is not discussed in detail in this work. The suction-based grasping mechanism is simulated using Stonefish's glue functionality [3], which enables the dynamic creation of a fixed joint between two bodies during runtime. Upon successful alignment between the end effector and the starfish, a glue constraint is activated to rigidly attach the starfish to the end effector, emulating the effect of a suction pump. This approach provides a reliable abstraction of suction attachment while allowing controlled activation and deactivation of the grasp within the simulation framework. This setup allows testing of the coordinated operation of the vehicle, manipulator, and suction end effector in a realistic deep-sea scenario without requiring physical deployment.

### 4.3. Results

The experimental results demonstrate the successful execution of a fully autonomous deep-sea object recovery mission in simulation. Figure 3 provides an overview of the HSV motion throughout the mission, including both the three-dimensional INS-estimated trajectory and the corresponding position error evolution. The vehicle descends from the sea surface to a depth of 6,000 m; for clarity, only the portion of the trajectory below 5,990 m is shown. After reaching the seafloor, the vehicle executes a structured lawnmower-style coverage pattern, maintaining stable depth and heading throughout the survey phase. The INS-estimated trajectory is shown together with the commanded waypoints, illustrating the vehicle's ability to follow the planned path during the seafloor survey. The magnitude of the associated

---

[3] https://stonefish.readthedocs.io/en/latest/special.html#glue

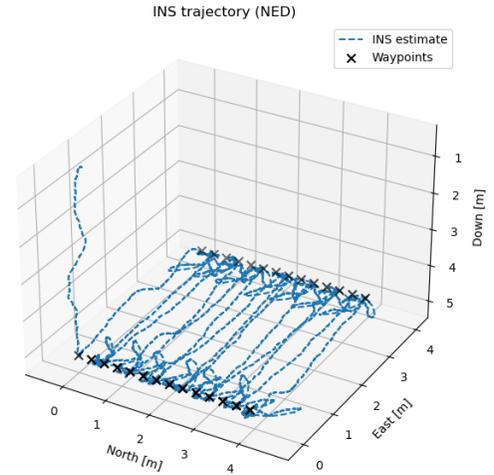

**FIGURE 3**: The vehicle descends from the sea surface to a depth of 6,000 m; for clarity, only the portion of the trajectory below 5,995 m is shown. After reaching the seafloor, the vehicle executes a lawnmower-style coverage pattern. The three-dimensional INS-estimated trajectory in the local NED frame is shown together with the commanded waypoints

INS position error with respect to the reference odometry, also shown in Fig. 3, highlights the gradual accumulation of navigation error over the duration of the mission, as expected for inertial navigation in the absence of external position updates. A more detailed comparison of the navigation performance is presented in Fig. 4, which shows the North, East, and Down position components estimated by the INS alongside the reference odometry. This per-axis comparison reveals axis-specific biases and drift behavior, providing insight into the directional characteristics of the INS error during long-duration operation. During the seafloor survey, the down-looking camera and onboard illumination enabled reliable detection of the target starfish. Upon detection, the vehicle transitioned from survey motion to a stabilized hovering state above the target. The manipulator was then deployed, and the inverse kinematics controller with acceleration feed-forward guided the suction cup to a precise pre-grasp pose above the starfish. Successful activation of the suction pump resulted in secure attachment of the starfish to the end effector, confirming correct coordination between perception, vehicle stabilization, manipulator control, and end effector actuation. Figure 5 illustrates the sequential stages of detection, stabilization, and grasp execution. The results validate the feasibility of the proposed control architecture for autonomous object recovery in hadal-like conditions. The experiments highlight the importance of tightly integrated vehicle and manipulator control, as well as realistic sensing and dynamics simulation, to develop robust deep-sea intervention behaviors.

### 5. CONCLUSION

We have presented a simulation-based investigation of autonomous object recovery in the hadal zone using a vehicle with integrated manipulation capability. Unlike previous studies that focus on shallow or mid-water environments, this work demon-strates realistic simulation of full-ocean-depth missions, capturing depth-dependent changes in vehicle dynamics, mass distribution, and volume under extreme pressure, along with their impact on onboard sensors. Using the enhanced Stonefish simulator, which incorporates these effects, as well as realistic hydrodynamics and environmental interactions, we show that coordinated control strategies can achieve reliable navigation and manipulation in challenging subsea conditions while reducing risk and cost prior to field deployment. Beyond control validation, high-fidelity simulation provides a natural foundation for higher-level autonomy and reasoning frameworks, as demonstrated in recent work on shared and multi-agent underwater autonomy using knowledge graphs and retrieval-augmented generation [23], as well as distributed cognitive architectures for underwater robots [24]. For long-duration and large-area missions, such frameworks can further benefit from efficient environment representation and mapping solutions, such as GPU-accelerated occupancy grids for real-time SLAM and persistent world modeling [25]. Together, these developments position simulation-based pipelines, particularly those capable of capturing realistic hadal-depth physics, as an important enabler for scalable, intelligent, and autonomous deep-sea robotic intervention.

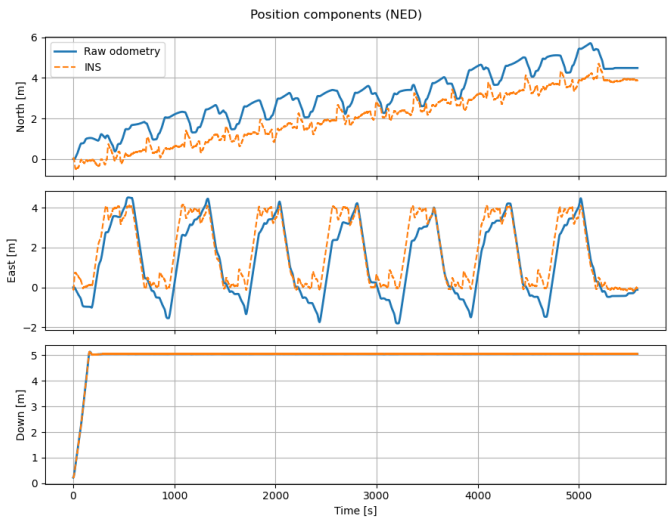

**FIGURE 4**: Per-axis position comparison between INS and reference odometry. North, East, and Down position components estimated by the INS are compared against the reference odometry as functions of time, highlighting axis-specific biases and drift behavior.

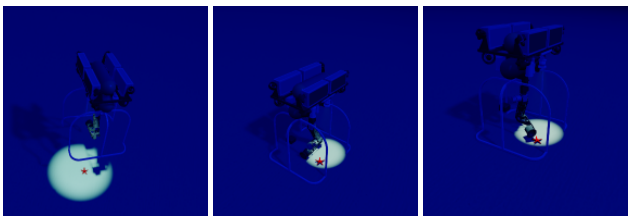

**FIGURE 5**: Sequential stages of the autonomous grasping operation. (Left) Visual detection and localization of the starfish by the perception system. (Center) Vehicle stabilization and landing above the target zone. (Right) Manipulator motion and suction-based grasp of the starfish.

## REFERENCES

[1] Weston, Johanna N J and Jamieson, Alan J. "Exponential growth of hadal science: perspectives and future directions identified using topic modelling." *ICES Journal of Marine Science* Vol. 79 No. 4 (2022): pp. 1048–1062. DOI 10.1093/icesjms/fsac074. URL https://academic.oup.com/icesjms/article-pdf/79/4/1048/43782814/fsac074.pdf, URL https://doi.org/10.1093/icesjms/fsac074.

[2] Maeda, Yosaku, Kakami, Hitoshi, Watanabe, Yoshitaka, Aso, Tatsuya, Iwashita, Kazuya, Kaneko, Tatsuya, Inoue, Tomoya, Koike, Hiroaki, Noborio, Yuhei and Maki, Atuso. "Towards an Autonomous Sampling System for the Hadal Zone." *2025 IEEE Underwater Technology (UT)*: pp. 1–5. 2025. DOI 10.1109/UT61067.2025.10947461.

[3] Maeda, Yosaku, Kakami, Hitoshi and Aso, Tatsuya. "Development of an Electric Robotic Arm for the Hadal Zone." *OCEANS 2024 - Halifax*: pp. 1–5. 2024. DOI 10.1109/OCEANS55160.2024.10753793.

[4] Bowen, Andrew D., Yoerger, Dana R., Taylor, Chris, McCabe, Robert, Howland, Jonathan, Gomez-Ibanez, Daniel, Kinsey, James C., Heintz, Matthew, McDonald, Glenn, Peters, Donald B., Fletcher, Barbara, Young, Chris, Buescher, James, Whitcomb, Louis L., Martin, Stephen C., Webster, Sarah E. and Jakuba, Michael V. "The Nereus hybrid underwater robotic vehicle for global ocean science operations to 11,000m depth." *OCEANS 2008*: pp. 1–10. 2008. DOI 10.1109/OCEANS.2008.5151993.

[5] Li, Jixu, Tang, Yuangui, Wang, Jian, Chen, Cong, Li, Yiping, Yan, Shuxue and Tian, Qiyan. "The Application of the Haidou Autonomous and Remotely-operated Vehicle in the third Mariana Trench scientific expedition of China.": pp. 1–4. 2019. DOI 10.23919/OCEANS40490.2019.8962660.

[6] Antonelli, Gianluca. *Underwater Robots - 2nd Edition - Motion and Force Control of Vehicle-Manipulator Systems*. Vol. 2 (2006). DOI 10.1007/11540199.

[7] Leonessa, Alexander. "Underwater Robots: Motion and Force Control of Vehicle-Manipulator Systems (G. Antonelli; 2006) [Book Review]." *IEEE Control Systems Magazine* Vol. 28 No. 5 (2008): pp. 138–139. DOI 10.1109/MCS.2008.927329.

[8] Wang, Zhenhua and Cui, Weicheng. "For safe and compliant interaction: an outlook of soft underwater manipulators." *Proceedings of the Institution of Mechanical Engineers, Part M: Journal of Engineering for the Maritime Environment* Vol. 235 (2020): p. 147509022095091. DOI 10.1177/1475090220950911.

[9] Zhang, Yunce, He, Weidong and Wang, Tao. "Design, modeling, and control of underwater stiffness-enhanced flexible manipulator." *Ocean Engineering* Vol. 308 (2024): p. 118302. DOI 10.1016/j.oceaneng.2024.118302.

[10] Mazzeo, Angela, Aguzzi, Jacopo, Calisti, Marcello, Canese, Simonepietro, Angiolillo, Michela, Allcock, A. Louise, Vecchi, Fabrizio, Stefanni, Sergio and Controzzi, Marco.

"Marine Robotics for Deep-Sea Specimen Collection: A Taxonomy of Underwater Manipulative Actions." *Sensors* Vol. 22 No. 4 (2022). DOI 10.3390/s22041471. URL https://www.mdpi.com/1424-8220/22/4/1471.

[11] Grimaldi, M., Nakath, D., She, M. and Köser, K. "INVESTIGATION OF THE CHALLENGES OF UNDERWATER-VISUAL-MONOCULAR-SLAM." *ISPRS Annals of the Photogrammetry, Remote Sensing and Spatial Information Sciences* Vol. X-1/W1-2023 (2023): p. 1113–1121. DOI 10.5194/isprs-annals-x-1-w1-2023-1113-2023. URL http://dx.doi.org/10.5194/isprs-annals-X-1-W1-2023-1113-2023.

[12] Kumamoto, Hikaru, Shirakura, Naoki, Takamatsu, Jun and Ogasawara, Tsukasa. "Underwater Suction Gripper for Object Manipulation with an Underwater Robot.": pp. 1–7. 2021. DOI 10.1109/ICM46511.2021.9385703.

[13] Cieślak, Patryk. "Stonefish: An Advanced Open-Source Simulation Tool Designed for Marine Robotics, With a ROS Interface." *OCEANS 2019 - Marseille*. 2019. DOI 10.1109/OCEANSE.2019.8867434.

[14] Grimaldi, Michele, Cieslak, Patryk, Ochoa, Eduardo, Bharti, Vibhav, Rajani, Hayat, Carlucho, Ignacio, Koskinopoulou, Maria, Petillot, Yvan R. and Gracias, Nuno. "Stonefish: Supporting Machine Learning Research in Marine Robotics." (2025). URL 2502.11887, URL https://arxiv.org/abs/2502.11887.

[15] Zhang, Mabel M., Choi, Woen-Sug, Herman, Jessica, Davis, Duane, Vogt, Carson, McCarrin, Michael, Vijay, Yadunund, Dutia, Dharini, Lew, William, Peters, Steven and Bingham, Brian. "DAVE Aquatic Virtual Environment: Toward a General Underwater Robotics Simulator." (2022). URL 2209.02862, URL https://arxiv.org/abs/2209.02862.

[16] Potokar, Easton, Lay, Kalliyan, Norman, Kalin, Benham, Derek, Ashford, Spencer, Peirce, Randy, Neilsen, Tracianne B., Kaess, Michael and Mangelson, Joshua G. "HoloOcean: A Full-Featured Marine Robotics Simulator for Perception and Autonomy." *IEEE Journal of Oceanic Engineering* Vol. 49 No. 4 (2024): pp. 1322–1336. DOI 10.1109/JOE.2024.3410290.

[17] Romrell, Blake, Austin, Abigail, Meyers, Braden, Anderson, Ryan, Noh, Carter and Mangelson, Joshua G. "A Preview of HoloOcean 2.0." (2025). URL 2510.06160, URL https://arxiv.org/abs/2510.06160.

[18] Song, Jingyu, Ma, Haoyu, Bagoren, Onur, Sethuraman, Advaith V., Zhang, Yiting and Skinner, Katherine A. "OceanSim: A GPU-Accelerated Underwater Robot Perception Simulation Framework." (2025). URL 2503.01074, URL https://arxiv.org/abs/2503.01074.

[19] Aldhaheri, Sara, Hu, Yang, Xie, Yongchang, Wu, Peng, Kanoulas, Dimitrios and Liu, Yuanchang. "Underwater Robotic Simulators Review for Autonomous System Development." (2025). URL 2504.06245, URL https://arxiv.org/abs/2504.06245.

[20] Grimaldi, Michele, Alkaabi, Nouf, Ruscio, Francesco, Rua, Sebastian Realpe, Garcia, Rafael and Gracias, Nuno. "Realtime Seafloor Segmentation and Mapping." (2025). URL 2504.10750, URL https://arxiv.org/abs/2504.10750.

[21] Lu, Daohua, Ning, Yichen, Wang, Jia, Du, Kaijie and Song, Cancan. "Research on Model Reduction of AUV Underwater Support Platform Based on Digital Twin." *Journal of Marine Science and Engineering* Vol. 12 No. 9 (2024). DOI 10.3390/jmse12091673. URL https://www.mdpi.com/2077-1312/12/9/1673.

[22] Adetunji, Favour O., Ellis, Niamh, Koskinopoulou, Maria, Carlucho, Ignacio and Petillot, Yvan R. "Digital Twins Below the Surface: Enhancing Underwater Teleoperation." (2024). URL 2402.07556, URL https://arxiv.org/abs/2402.07556.

[23] Grimaldi, Michele, Cernicchiaro, Carlo, Rua, Sebastian Realpe, El-Masri-El-Chaarani, Alaaeddine, Buchholz, Markus, Michael, Loizos, Rodriguez, Pere Ridao, Carlucho, Ignacio and Petillot, Yvan R. "Advancing Shared and Multi-Agent Autonomy in Underwater Missions: Integrating Knowledge Graphs and Retrieval-Augmented Generation." (2025). URL 2507.20370, URL https://arxiv.org/abs/2507.20370.

[24] Buchholz, Markus, Carlucho, Ignacio, Grimaldi, Michele and Petillot, Yvan R. "Distributed AI Agents for Cognitive Underwater Robot Autonomy." (2025). URL 2507.23735, URL https://arxiv.org/abs/2507.23735.

[25] Grimaldi, Michele, Palomeras, Narcis, Carlucho, Ignacio, Petillot, Yvan R and Rodriguez, Pere Ridao. "FRAGG-Map: Frustum Accelerated GPU-Based Grid Map." *2024 IEEE/RSJ International Conference on Intelligent Robots and Systems (IROS)*: pp. 1138–1144. 2024. IEEE.